\documentclass{article}
\usepackage{spconf,amsmath,graphicx}
\usepackage{multirow}
\usepackage{booktabs}
\usepackage{caption}
\usepackage{subcaption}
\usepackage{graphicx}
\usepackage{xcolor}
\usepackage{amssymb}
\usepackage[noadjust]{cite}
\usepackage{comment}
\usepackage{dsfont}
\usepackage{amsmath}
\usepackage{float}
\usepackage{bm}
\usepackage{hyperref}
\usepackage{multicol}
\usepackage{multirow}
\usepackage{amssymb}
\usepackage[normalem]{ulem}

\DeclareMathOperator*{\argmin}{arg\,min}

\title{A Facial Affect Analysis System for Autism Spectrum Disorder}

\name{
Beibin Li$^{\star \dagger}$ ~\quad
Sachin Mehta$^\star$ ~\quad
Deepali Aneja$^\star$ ~\quad
Claire Foster$^\ddagger$ \\
Pamela Ventola$^\mathsection$ ~\quad
Frederick Shic$^{\star \dagger}$ ~\quad
Linda Shapiro$^\star$
\vspace{4mm}
}

\address{
$^{\star}$ Paul G. Allen School of Computer Science and Engineering, University of Washington, Seattle, WA ~\\
$^{\dagger}$ Seattle Children's Research Institute, Seattle, WA ~\\
$^{\ddagger}$ Department of Psychology, Binghamton University, Binghamton, NY ~\\
$^{\mathsection}$ Yale Child Study Center, School of Medicine, Yale University, New Haven, CT
}

\begin{document}
\maketitle

%\ninept
%

%
\begin{abstract}
In this paper, we introduce an end-to-end machine learning-based system for classifying autism spectrum disorder (ASD) using facial attributes such as expressions, action units, arousal, and valence. Our system classifies ASD using representations of different facial attributes from convolutional neural networks, which are trained on \textit{images in the wild}. Our experimental results show that different facial attributes used in our system are statistically significant and improve sensitivity, specificity, and F1 score of ASD classification by a large margin. In particular, the addition of different facial attributes improves the performance of ASD classification by about 7\% which achieves a F1 score of 76\%.

\end{abstract}
\begin{keywords}
Affective Computing, Convolutional Neural Networks, Healthcare, Autism Spectral Disorder
\end{keywords}

\section{Introduction}
\label{sec:intro}

Autism spectrum disorder (ASD) is a neurodevelopment disorder that affects social communication and behavior of children \cite{wang2004neural,loth2018facial}. According to the Centers for Disease Control and Prevention, one out of 59 children is diagnosed with ASD in the United States\footnote{\url{https://www.cdc.gov/ncbddd/autism/data.html}}. Diagnosing ASD can be difficult because (1) the type and severity of symptoms have a wide spectrum, and (2) the behavior of children with autism is dependent on non-autism-specific factors such as cognitive functioning and age \cite{charman2014variability}. Facial attributes including expressions have been suggested as effective markers in autism related clinical studies \cite{wang2004neural,loth2018facial,dapretto2006understanding,ozgen2013predictive}.

Convolutional neural networks (CNNs) produce state-of-the-art results for recognizing different facial attributes (e.g., expressions, gender, and action units (AUs)) in the wild \cite{mollahosseini2017affectnet,zeng2018facial,li2018patch,kervadec2018cake,benitez2017emotionet}. The high accuracy achieved in these recognition tasks can be attributed to large-scale labeled datasets, such as AffectNet \cite{mollahosseini2017affectnet} and EmotioNet \cite{fabian2016emotionet}, that enable CNNs to learn rich and generalizable representations. However, datasets at such scale do not exist for ASD, making it difficult to apply CNNs directly in the autism field.

In this paper, we introduce a system for ASD classification using facial attributes. Along with two widely used categorical facial attributes (facial expressions and AUs) for natural images, our system also predicts two continuous facial \textit{affect} attributes (arousal and valence) that have been found to be effective in autism related clinical studies \cite{rudovic2018personalized}. For simplicity, we use facial attributes to represent \textit{facial expression}, \textit{AUs}, \textit{arousal}, and \textit{valence}. Since there are no publicly available datasets for autism with \textit{all} of these different attributes, we learn representations for these attributes by leveraging two large-scale facial datasets of natural images that are collected in a wide variety of settings, including age, gender, race, pose, and lighting variations. The contributions of this work are: (1) present an ASD classification system based on facial attributes, (2) show the importance of these facial attributes in improving the performance of our system through statistical analysis, and (3) analysis of single vs. multi-task learning for facial attribute recognition.

% Our experimental results suggest that representations of different facial attributes improve the performance of our system by about 7\%. Our statistical analysis also proves the importance of these attributes.

\section{Related Work}
\label{sec:related}

We train a CNN-based model that takes a facial image as 
input and outputs four facial attributes to be used for ASD prediction. In this section, we briefly review the existing work for facial attribute recognition and their application in autism.

\noindent \textbf{Facial attribute recognition:} With recently curated, large-scale datasets \cite{mollahosseini2017affectnet,fabian2016emotionet,barsoum2016training,calvo2008facial,zhang2014bp4d}, it has become possible to train CNNs for facial attribute recognition. These networks can learn facial representations either independently \cite{zeng2018facial,wiles2018self,li2018deep} or simultaneously \cite{hu2018deep}. Most existing datasets contain annotations for one or two facial attributes. In this work, we combine two large-scale datasets \cite{mollahosseini2017affectnet,fabian2016emotionet} and train a model to produce four facial attributes simultaneously for ASD classification.

\begin{figure*}[t!]
    \centering
    \includegraphics[width=2\columnwidth]{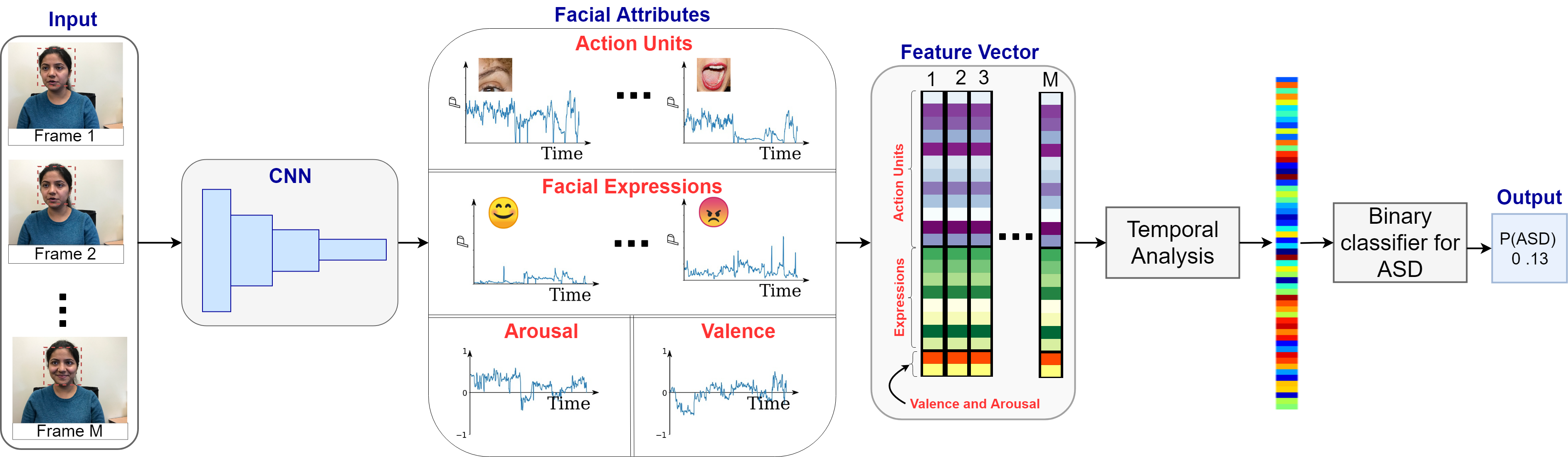}
    \caption{Overview of our end-to-end system for autism spectrum disorder (ASD) classification using facial attributes.}
    \label{fig:sys}
\end{figure*}

\noindent \textbf{Facial attributes for autism}: Various clinical studies have shown that facial attributes, including expressions \cite{wang2004neural,loth2018facial}, emotions \cite{dapretto2006understanding}, and morphological features \cite{ozgen2013predictive}, are effective markers for autism. With recent developments in technology, including sensors and artificial intelligence, affective computing is gaining interest in the autism community. Egger et al. \cite{egger2018automatic} use head orientation and expression to study autism-related behavior. Rudovic et al. \cite{rudovic2018personalized} use facial landmarks and body pose along with captured audio and bio-signals for an automatic perception of children's affective state and engagement. In this work, we use representations of different facial attributes for ASD classification. 

\section{Our System for ASD classification}
\label{sec:system}
Our system, shown in Fig. \ref{fig:sys}, takes a video as an input and uses CNN to extract four facial attributes per frame: facial expressions, AUs, arousal, and valence on the participant's detected face\footnote{We use a HoG-based face detector for its good trade-off between speed and accuracy on an iPad. However, any other face detector can be used.}. These outputs corresponding to four facial attributes are concatenated to form a $k$-dimensional feature vector per frame, represented as $\bm{f}^t = \{\bm{f}_{au}^t, \bm{f}_{expr}^t, f_{aro}^t, f_{val}^t \} \in \mathbb{R}^k$ where $\bm{f}_{au} \in \mathbb{R}^n$ and $\bm{f}_{expr} \in \mathbb{R}^m$ are feature vectors corresponding to $n$ AUs and $m$ expressions available in the dataset, while $f_{aro}$ and $f_{val}$ are scalar values between -1 and 1 that correspond to arousal and valence, respectively. We  apply temporal feature extraction methods on each vector $\bm{f}^t \in \mathbb{R}^k$ to extract a single lower-dimensional temporal feature vector $\bm{\hat{f}} \in \mathbb{R}^l$ per video. Each temporal feature vector $\bm{\hat{f}}$ is fed to a binary classifier for ASD prediction. In this section, we describe our system for ASD classification in detail.

\noindent \textbf{Facial attribute recognition:} In this paper, we are interested in ASD classification. However, there are no large publicly available datasets that provide annotated videos with facial attributes as labels for ASD. 
Therefore, we use publicly available large-scale datasets that provide one or more facial attributes for natural images in the wild. We use these datasets to train a CNN-based model that simultaneously predicts different facial attributes. Our network is a standard CNN that learns spatial representations by stacking convolution and down-sampling units, as shown in Table \ref{tab:structure}. During training, we minimize the following multi-task loss function \cite{evgeniou2004regularized}:
\begin{equation}
    \argmin \sum_{t=1}^{T} \sum_{i=1}^{N} l(y^t_i, f(\bm{x}^t_i; \bm{w}^t)) + \phi(\bm{w}^t)
\label{eq:MTLLoss}
\end{equation}
Here, $f(\bm{x}^i; \bm{w}^t)$ is a function of input $\bm{x}^t$ and learnable parameters $\bm{w}^t$, $l(.)$ is a task specific loss function, $\phi(\bm{w}^t)$ is the regularization term, $T$ is the number of tasks, and $N$ is the number of data samples. 

\begin{table}[b!]
\centering
\resizebox{0.75\columnwidth}{!}{
    \begin{tabular}{l|c|c|c}
        \toprule
        {\bf Layer/} & \multirow{2}{*}{\bf Repeat}  & \multicolumn{2}{c}{\bf Output} \\
        \cline{3-4}
         {\bf Stride} &   & {\bf Size} & {\bf Channels} \\
        \midrule
        Conv-3/2 & 1 & $112 \times 112$ & 32 \\
        \hline
        CU/2 & 1 & $56 \times 56$  & 32 \\
        CU/1 & 1 & $56 \times 56$ & 32 \\
        \hline
        CU/2 & 1 & $28 \times 28$ & 64 \\
        CU  & 3 & $28 \times 28$ & 64 \\
        \hline
        CU/2& 1 & $14 \times 14$  & 128 \\
        CU & 7 & $14 \times 14$ & 128 \\
        \hline
        CU/2 & 1 & $7 \times 7$ & 256 \\
        CU/1  & 3 & $7 \times 7$ & 256 \\
        DWConv-3/1 & 1 & $7 \times 7$ & 512 \\
        \hline
        Avg. pool & \multicolumn{1}{c|}{ } & $1\times 1$  & 512 \\
        \hline
        Linear $\times 4$ & & \multicolumn{2}{c}{ $C_{expr}, C_{au}, C_{val}, C_{aro}$ }  \\
        \bottomrule
    \end{tabular}
}
\caption{Overview of our CNN architecture. Four linear layers are used as last layer in parallel; each branch predicting $C_{expr}$ expressions, $C_{au}$ AUs,  $C_{aro}$ arousal value, and $C_{val}$ valence value. We note that $C_{aro}$ and $C_{val}$ are between -1 and 1. Here, Conv-3 and DWConv-3 represents $3\times3$ standard and depth-wise convolutional layers, and CU represents convolutional unit. We use three different CUs (BottleNeck \cite{he2016deep},  MobileNet \cite{howard2017mobilenets}, and EESP \cite{mehta2018espnetv2}) in our study.}
\label{tab:structure}
\end{table}

\noindent \textbf{ASD classification:} After training our model on the publicly available facial datasets, we generate a $k$-dimensional feature vector $\bm{f}^t$ for each frame in the participant video by feeding the video into our trained CNN-model frame by frame. These vectors are concatenated to form a feature matrix $\bm{F} \in \mathbb{R}^{M \times k}$ per video, where $M$ denotes the total number of frames in the video. Due to the temporal nature of the data, there may be redundancies in the feature matrix $\bm{F}$ that could hinder the analysis of the differences between ASD and non-ASD participants. Therefore, we project this high-dimensional feature matrix $\bm{F}$ to an $l$-dimensional vector $\bm{\hat{f}} \in \mathbb{R}^{l}$ using temporal analysis methods. In particular, we compute mean vector $\bm{m} \in \mathbb{R}^{k}$ and standard deviation vector $\bm{\sigma} \in \mathbb{R}^{k}$ that contain the mean and standard deviation values across our $k$ features.

In addition,  we compute an activation vector $\bm{a} \in \mathbb{R}^{n}$ that captures the mean activation time per action unit, because of its significance in interpretability.
%\sout{the facial action unit's activation time is useful for interpretability.} 
We define $\bm{a}$ as: $\bm{a}(i) = \cfrac{1}{M} \sum\limits_{t=0}^M \mathds{1}_{ f_{au}^t(i) > \tau}$ where $\mathds{1}$ is an indicator function and $\tau$ is a threshold. We use $\tau=0.5$ in our experiments. 

% v0
% After training our model on the publicly available facial datasets, we feed every frame of the participant video to our trained CNN-model and generate a $k$-dimensional feature vector $\bm{f}^t$ per frame. 

Similarly, because it has been shown that the percentages of positive arousal $p_{aro}$ and positive valence $p_{val}$ frames are meaningful for autism related studies \cite{egger2018automatic,rudovic2018personalized}, we also compute these features. We concatenate the vectors and scalars obtained after temporal analysis to produce $l$-dimensional feature vector $\bm{\hat{f}} = \{\bm{m}, \bm{\sigma}, \bm{a}, p_{aro}, p_{val}\}$. We feed $\bm{\hat{f}}$ to a binary classifier to predict if the participant is affected by ASD or not.
% v0

\section{Experiments}
In this section, we first study the performance of our system on facial attribute recognition on different facial datasets. We then study the impact of each facial attribute on ASD classification along with their statistical significance. 
% v0

\subsection{Facial attribute recognition}
\noindent \textbf{Dataset:}  Most of the existing datasets provide annotations for one or two facial attributes. To train a network with all four facial attributes (expressions, AUs, arousal, and valence), we combine two publicly available datasets\footnote{All images do not have labels for all facial attributes. Therefore, we fill the missing attribute value with an UNK which is ignored during training. The expression, arousal, and valence labels are from AffectNet, and the AU labels are from EmotioNet.}: (1) AffectNet \cite{mollahosseini2017affectnet} and (2) EmotioNet \cite{fabian2016emotionet}. The resulting dataset contains about 1.2 million samples. For AffectNet, we split the training set into two subsets: training (285K) and validation (2.4K). Following \cite{zeng2018facial}, we use AffectNet's validation set as the test set (5.5K). For EmotioNet, we split the training set into three subsets: training (754K), validation (63K), and testing (126K). 

\noindent \textbf{Training details:} We train our models in PyTorch for a total of 30 epochs using Stochastic Gradient Descent with a momentum of 0.9 and an initial learning rate of 0.01. For faster convergence, we decrease the learning rate by 5\% after every epoch. Annotations for facial attributes are different: some are continuous (arousal and valence), and some are discrete (AUs and expressions). Therefore, we use task-specific loss functions to learn representations for different facial attributes. In particular, we minimize cross-entropy loss for expression, binary cross-entropy loss for AUs, and sum of L1 and L2 loss for arousal and valence respectively. For multi-task learning, we use the sum of task-specific loss functions, similar to \cite{evgeniou2004regularized, zhang2014facial}. We also use inverse class probability weighting scheme for each loss function to address the class imbalance. We use standard data augmentation strategies such as random flipping, cropping, rotation, and shearing while training our models.

\begin{figure}[t!]
    \centering
    \begin{subfigure}[b]{0.7\columnwidth}
        \centering
        \includegraphics[width=\columnwidth]{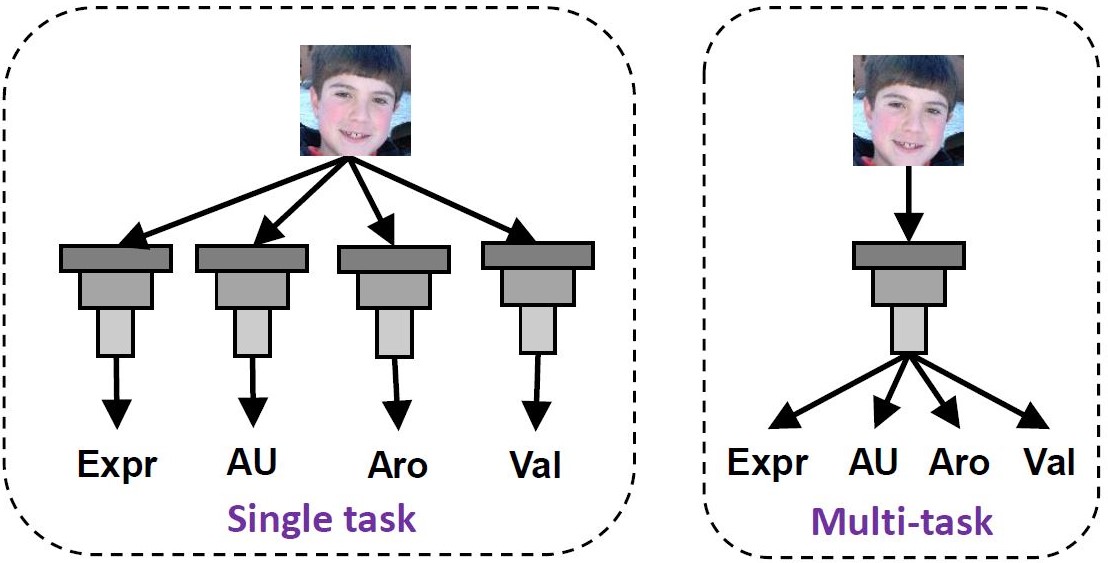}
        \caption{ }
        \label{fig:singleMultiTask}
    \end{subfigure}%
    \vfill
    \begin{subfigure}[b]{\columnwidth}
    \centering
    \resizebox{0.9\columnwidth}{!}{
        \begin{tabular}{l|rr|cccc}
        \toprule
         \multirow{2}{*}{{\bf CNN Unit}} & \multirow{2}{*}{{\bf \# Params}} & \multirow{2}{*}{{\bf FLOPs}} & {\bf Expr} & {\bf AU} & {\bf Val} & {\bf Aro}\\
         &  & & (F1) & (mF1Acc) & (CC) & (CC) \\
        \midrule
         \multicolumn{7}{c}{\it Single-task} \\
         \midrule
        Bottleneck \cite{he2016deep}          & 25.9 M & 3.4 B & 0.56 & {\bf 0.78} & 0.63 & 0.54 \\
        MobileNet \cite{howard2017mobilenets} & 24.8 M &  3.1 B  & 0.57 & 0.77 & 0.64 & 0.52 \\
        EESP \cite{mehta2018espnetv2}         & 9.7 M & 1.2 B & 0.57 & 0.76 & 0.64 & 0.52 \\
         \midrule
         \multicolumn{7}{c}{\it Multi-task} \\
         \midrule
         Bottleneck \cite{he2016deep}          & 6.5 M & 0.85 B & {\bf 0.58} & 0.75 & 0.68 & 0.61 \\
         MobileNet \cite{howard2017mobilenets} & 6.2 M & 0.78 B & {\bf 0.58} & 0.75 & 0.68 & {\bf 0.62} \\
          EESP \cite{mehta2018espnetv2}        & \textbf{2.4 M} & \textbf{0.29 B} & {\bf 0.58} & 0.75 & {\bf 0.69} & 0.61 \\
         \bottomrule
        \end{tabular}
        }
    \caption{ }
    \label{fig:singVsMulti}
    \end{subfigure}
    \caption{Single vs. multi-task learning: (a) comparison between single and multi-task learning, and (b) qualitative performance of single and multi-task learning models. Here, Expr, AU, Val, Aro, and FLOPs indicate expression, AUs,  arousal, valence, and floating points operations  respectively.} 
\end{figure}

\noindent \textbf{Results:} We use CNNs to predict facial attributes for a given input image in both single-task and multi-task settings. In the single-task set-up, the input image was fed to four \textit{different} CNNs, where each CNN predicts a different facial attribute. In the multi-task set-up, the input image was fed to a \textit{single} CNN that predicts all facial attributes at once. A comparison between single and multi-task learning set-up is shown in Fig. \ref{fig:singleMultiTask}. Furthermore, different CNN units (e.g. bottleneck block in ResNet \cite{he2016deep}) have been proposed in the literature to learn richer representations. To find a suitable trade-off between accuracy and a network's complexity, we study three different convolutional units: (1) the Bottleneck unit \cite{he2016deep}, (2) the EESP unit \cite{mehta2018espnetv2}, and (3) the MobileNet unit \cite{howard2017mobilenets}.  Following the conventions in the literature, we use the following metrics to evaluate the performance of our model: (1) an average of F1-score and accuracy for AUs \cite{fabian2016emotionet}, (2) F1-score for expressions \cite{mollahosseini2017affectnet}, and (3) correlation coefficient (CC) for  arousal and valence \cite{mollahosseini2017affectnet}. 

We make the following observations from the results shown in Fig.\ref{fig:singVsMulti} : (1) multi-task learning delivers better performance than single-task learning for all different facial attributes except AUs. In particular, the multi-task learning-based system outperforms the single-task learning-based system for arousal by about $8\%$, and (2) the EESP unit delivers similar performance to the Bottleneck and the MobileNet units, but is much more efficient and uses much fewer parameters and floating point operations (FLOPs). The second observation is in contrast to other large scale datasets, such as the ImageNet, where the complex models deliver better performance. This suggests that facial expression datasets are not as complex as the ImageNet and that complex CNN models (e.g., \cite{he2016deep}) learn redundant parameters without giving significant performance gains. We note that the recognition performance of our method is on par with existing CNN-based methods \cite{mollahosseini2017affectnet,zeng2018facial}.

\subsection{Application to ASD classification}

\noindent \textbf{Dataset:} We collect a video dataset of 105 children (ASD: 62 and non-ASD: 43) with one video per participant using an iPad application; 88 of these children (ASD: 49 and non-ASD: 39) finish the experiment and then consent to use their data for our research. The diagnostic labels, \textit{ASD} or \textit{non-ASD}, are provided by clinicians based on the neuropsychological tests, which are done independently of these experiments.

During the experiment, each participant watches an expert-designed video stimulus on an iPad. The video stimulus is a compilation of short video clips that display both dynamic naturalistic scene and social communication scene together. These clips are shown simultaneously, side-by-side, on a vertically split iPad screen. While the participant watches a video, our application captures and records the participant's facial response using the iPad's front camera. The video recorded using the iPad application is about 6 minutes and 35 seconds (9,575 valid frames) per participant.

\noindent \textbf{Methods:} We construct a 22-dimensional feature vector from four facial attributes produced by the CNN.
The first 12 values in this vector represent the probability of each action unit, the next 8 values represent the probability of each expression, and the last two values represent the arousal and valence attributes. This results in a $9575 \times 22$-dimensional matrix per participant.  We then use temporal analysis methods (see Section \ref{sec:system}) to construct a 58-dimensional feature vector per participant\footnote{Temporal features can also be learned using methods such as RNNs and temporal CNNs. However, we find these methods exhibit poor generalizability on our dataset. This is likely because these methods require a large amount of training data.}.
This feature vector comprises 44 values of mean and standard deviation per dimension ($22 \times 2$), 12 values representing mean percentage activation time of action units, and two values representing the percentage of positive arousal and positive valence. We train seven binary classifiers (logistic regression, LASSO, LDA, QDA, SVM with RBF kernel, XGBoost, and two-hidden-layer neural network (NN)) using these 58-dimensional feature vectors for ASD classification. Since the dataset is limited, we measure the classification performance (F1 score, sensitivity, and specificity) using leave-one-out cross-validation.

\begin{figure}[t!]
    \centering
    \begin{subfigure}[b]{0.9\columnwidth}
    \centering
        \includegraphics[width=\columnwidth]{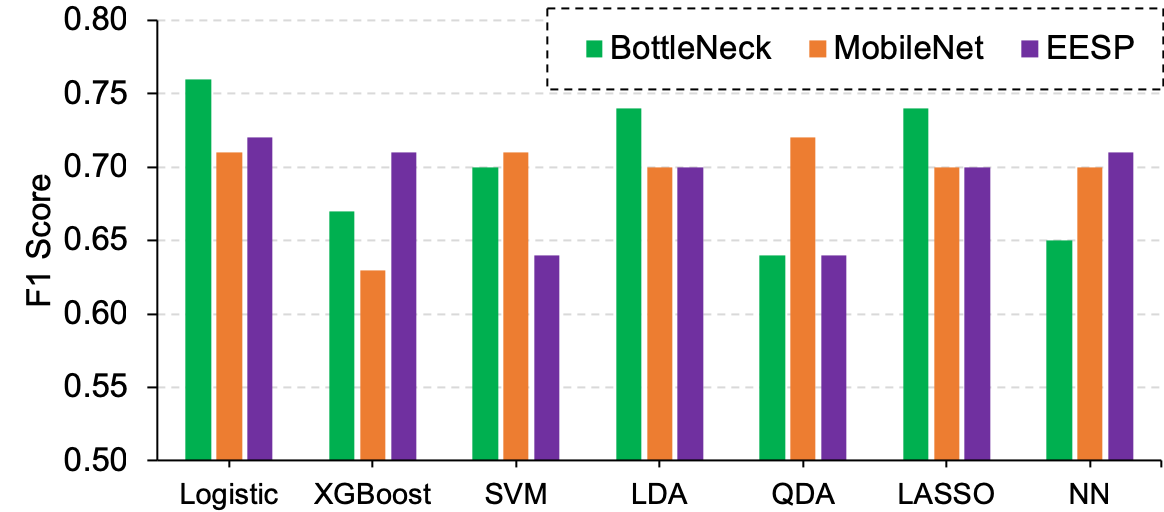}
        \caption{}
        \label{fig:algoCompare}
    \end{subfigure}
    \vfill
    \begin{subfigure}[b]{0.62\columnwidth}
    \resizebox{\columnwidth}{!}{
        \begin{tabular}{llll|ccc}
            \toprule
            \multicolumn{4}{c|}{\bf Facial attributes} & \multirow{2}{*}{\textbf{F1}} & \multirow{2}{*}{\textbf{Sensitivity}}   & \multirow{2}{*}{\textbf{Specificity}}   \\
            \cline{1-4}
            \multicolumn{1}{c}{\textbf{AU}} & \multicolumn{1}{c}{\textbf{Aro}} & \multicolumn{1}{c}{\textbf{Val}} & \multicolumn{1}{c|}{\textbf{Expr}} &  \\
            \midrule
            \checkmark &  &  &  & 0.69  & 0.69  & 0.62 \\
            \checkmark & \checkmark &  &  & 0.72  & 0.71    & 0.67 \\
            \checkmark & \checkmark & \checkmark &  & 0.69  & 0.67 & 0.67 \\
            \checkmark & \checkmark & \checkmark & \checkmark & \textbf{0.76} & \textbf{0.76} & \textbf{0.69} \\
            \bottomrule
        \end{tabular}
    }
        \caption{}
        \label{fig:impactFacAtt}
    \end{subfigure}
    \hfill
    \begin{subfigure}[b]{0.36\columnwidth}
        \centering
        \resizebox{0.9\columnwidth}{!}{
        \begin{tabular}{l|c}
            \toprule
            {\bf Facial attributes} & {\bf p-value} \\
            \midrule
            Action Units (AUs) &  0.223 \\
            % Action Units (Lower Face) &  0.003 \\
            Arousal  (Aro) &  0.007  \\
            Valence  (Val) &  0.001  \\
            Expression (Expr) & 0.006 \\
            \bottomrule
        \end{tabular}
        }
        \caption{}
        \label{fig:ttest_domain}
    \end{subfigure}
    \caption{ASD classification results: (a) comparison of different binary classification methods, (b) impact of different facial attributes on the classification performance with BottleNeck as a CNN unit, and (c) statistical significance using Student's t-test of different facial attributes.}
    \label{fig:binCompare}
\end{figure}

\noindent \textbf{Results:} Fig. \ref{fig:algoCompare} compares the performances of seven ASD classifiers that use representations from different CNNs. Our system achieves the best F1 score, sensitivity, and specificity with the Bottleneck as the base feature extractor.
We also note that the ASD classification performance improves by 7\% when we add features related to arousal, valence, and facial expressions. This result is consistent with our statistical analysis (Fig. \ref{fig:ttest_domain})  where we found these three attributes are the most significant.

% v0
% ur system achieved the best performance (in terms of F1 score, sensitivity, and specificity) with Bottleneck as the base feature extractor.

\section{Conclusion}
We presented an end-to-end system for ASD classification using different facial attributes: facial expressions, AUs, arousal, and valence. The multi-task learning approach used in our experiments is more effective to classify different facial attributes than the single-task approach. We also showed that representations of different facial attributes used in our study are statistically significant and improve the ASD classification performance by about 7\% with F1 score of $76\%$.
\\

\noindent {\footnotesize \textbf{Acknowledgement:} This work is supported by NIH awards K01 MH104739, R21 MH103550; the NSF Expedition in Socially Assistive Robotics \#1139078; and Simons Award \#383661. We would like to thank  Nicholas Nuechterlein, Erin Barney, James Snider, Minah Kim, Yeojin Amy Ahn, Madeline Aubertine, Kelsey Jackson,  Quan Wang, Adham Atyabi, and participants for data collection and participation in this work.}

\clearpage 

\small{
\bibliographystyle{IEEEbib}
\bibliography{references}

\begin{thebibliography}{10}

\bibitem{wang2004neural}
A~Ting Wang, Mirella Dapretto, Ahmad~R Hariri, Marian Sigman, and Susan~Y
  Bookheimer,
\newblock ``Neural correlates of facial affect processing in children and
  adolescents with autism spectrum disorder,''
\newblock {\em Journal of the American Academy of Child \& Adolescent
  Psychiatry}, vol. 43, no. 4, pp. 481--490, 2004.

\bibitem{loth2018facial}
E~Loth, L~Garrido, J~Ahmad, E~Watson, A~Duff, and B~Duchaine,
\newblock ``Facial expression recognition as a candidate marker for autism
  spectrum disorder: how frequent and severe are deficits?,''
\newblock {\em Molecular autism}, 2018.

\bibitem{charman2014variability}
Tony Charman,
\newblock ``Variability in neurodevelopmental disorders: evidence from autism
  spectrum disorders,''
\newblock in {\em Neurodevelopmental Disorders}. 2014.

\bibitem{dapretto2006understanding}
Mirella Dapretto, Mari~S Davies, Jennifer~H Pfeifer, Ashley~A Scott, Marian
  Sigman, Susan~Y Bookheimer, and Marco Iacoboni,
\newblock ``Understanding emotions in others: mirror neuron dysfunction in
  children with autism spectrum disorders,''
\newblock {\em Nature neuroscience}, 2006.

\bibitem{ozgen2013predictive}
H~Ozgen, GS~Hellemann, MV~De~Jonge, FA~Beemer, and H~van Engeland,
\newblock ``Predictive value of morphological features in patients with autism
  versus normal controls,''
\newblock {\em Journal of autism and developmental disorders}, 2013.

\bibitem{mollahosseini2017affectnet}
A.~Mollahosseini, B.~Hasani, and M.~H. Mahoor,
\newblock ``Affectnet: A database for facial expression, valence, and arousal
  computing in the wild,''
\newblock {\em IEEE Transactions on Affective Computing}, 2018.

\bibitem{zeng2018facial}
Jiabei Zeng, Shiguang Shan, and Xilin Chen,
\newblock ``Facial expression recognition with inconsistently annotated
  datasets,''
\newblock in {\em Proceedings of the European conference on computer vision
  (ECCV)}, 2018, pp. 222--37.

\bibitem{li2018patch}
Yong Li, Jiabei Zeng, Shiguang Shan, and Xilin Chen,
\newblock ``Patch-gated cnn for occlusion-aware facial expression
  recognition,''
\newblock in {\em 2018 24th International Conference on Pattern Recognition
  (ICPR)}. IEEE, 2018, pp. 2209--2214.

\bibitem{kervadec2018cake}
Corentin Kervadec, Valentin Vielzeuf, St{\'e}phane Pateux, Alexis Lechervy, and
  Fr{\'e}d{\'e}ric Jurie,
\newblock ``Cake: Compact and accurate k-dimensional representation of
  emotion,''
\newblock in {\em Image Analysis for Human Facial and Activity Recognition
  (BMVC Workshop)}, 2018.

\bibitem{benitez2017emotionet}
C~Fabian Benitez-Quiroz, Ramprakash Srinivasan, Qianli Feng, Yan Wang, and
  Aleix~M Martinez,
\newblock ``Emotionet challenge: Recognition of facial expressions of emotion
  in the wild,''
\newblock {\em arXiv preprint arXiv:1703.01210}, 2017.

\bibitem{fabian2016emotionet}
C~Fabian Benitez-Quiroz, Ramprakash Srinivasan, and Aleix~M Martinez,
\newblock ``Emotionet: An accurate, real-time algorithm for the automatic
  annotation of a million facial expressions in the wild,''
\newblock in {\em Proceedings of the IEEE Conference on Computer Vision and
  Pattern Recognition}, 2016, pp. 5562--5570.

\bibitem{rudovic2018personalized}
Ognjen Rudovic, Jaeryoung Lee, Miles Dai, Bjorn Schuller, and Rosalind Picard,
\newblock ``Personalized machine learning for robot perception of affect and
  engagement in autism therapy,''
\newblock {\em Science. 3. 10.1126/scirobotics.aao6760.}, 2018.

\bibitem{barsoum2016training}
Emad Barsoum, Cha Zhang, Cristian~Canton Ferrer, and Zhengyou Zhang,
\newblock ``Training deep networks for facial expression recognition with
  crowd-sourced label distribution,''
\newblock in {\em Proceedings of the 18th ACM International Conference on
  Multimodal Interaction}. ACM, 2016, pp. 279--283.

\bibitem{calvo2008facial}
Manuel~G Calvo and Daniel Lundqvist,
\newblock ``Facial expressions of emotion (kdef): Identification under
  different display-duration conditions,''
\newblock {\em Behavior research methods}, vol. 40, no. 1, pp. 109--115, 2008.

\bibitem{zhang2014bp4d}
Xing Zhang, Lijun Yin, Jeffrey~F Cohn, Shaun Canavan, Michael Reale, Andy
  Horowitz, Peng Liu, and Jeffrey~M Girard,
\newblock ``Bp4d-spontaneous: a high-resolution spontaneous 3d dynamic facial
  expression database,''
\newblock {\em Image and Vision Computing}, vol. 32, no. 10, pp. 692--706,
  2014.

\bibitem{wiles2018self}
Olivia Wiles, A~Koepke, and Andrew Zisserman,
\newblock ``Self-supervised learning of a facial attribute embedding from
  video,''
\newblock {\em arXiv preprint arXiv:1808.06882}, 2018.

\bibitem{li2018deep}
Shan Li and Weihong Deng,
\newblock ``Deep facial expression recognition: A survey,''
\newblock {\em arXiv preprint arXiv:1804.08348}, 2018.

\bibitem{hu2018deep}
Guosheng Hu, Li~Liu, Yang Yuan, Zehao Yu, Yang Hua, Zhihong Zhang, Fumin Shen,
  Ling Shao, Timothy Hospedales, Neil Robertson, et~al.,
\newblock ``Deep multi-task learning to recognise subtle facial expressions of
  mental states,''
\newblock in {\em Proceedings of the European Conference on Computer Vision
  (ECCV)}, 2018, pp. 103--119.

\bibitem{egger2018automatic}
Helen~L Egger, Geraldine Dawson, Jordan Hashemi, Kimberly~LH Carpenter, Steven
  Espinosa, Kathleen Campbell, Samuel Brotkin, Jana Schaich-Borg, Qiang Qiu,
  Mariano Tepper, et~al.,
\newblock ``Automatic emotion and attention analysis of young children at home:
  a researchkit autism feasibility study,''
\newblock {\em npj Digital Medicine}, vol. 1, no. 1, pp. 20, 2018.

\bibitem{evgeniou2004regularized}
Theodoros Evgeniou and Massimiliano Pontil,
\newblock ``Regularized multi--task learning,''
\newblock in {\em Proceedings of the tenth ACM SIGKDD international conference
  on Knowledge discovery and data mining}. ACM, 2004, pp. 109--117.

\bibitem{he2016deep}
Kaiming He, Xiangyu Zhang, Shaoqing Ren, and Jian Sun,
\newblock ``Deep residual learning for image recognition,''
\newblock in {\em Proceedings of the IEEE conference on computer vision and
  pattern recognition}, 2016, pp. 770--778.

\bibitem{howard2017mobilenets}
Andrew~G Howard, Menglong Zhu, Bo~Chen, Dmitry Kalenichenko, Weijun Wang,
  Tobias Weyand, Marco Andreetto, and Hartwig Adam,
\newblock ``Mobilenets: Efficient convolutional neural networks for mobile
  vision applications,''
\newblock {\em arXiv preprint arXiv:1704.04861}, 2017.

\bibitem{mehta2018espnetv2}
Sachin Mehta, Mohammad Rastegari, Linda Shapiro, and Hannaneh Hajishirzi,
\newblock ``Espnetv2: A light-weight, power efficient, and general purpose
  convolutional neural network,''
\newblock {\em arXiv preprint arXiv:1811.11431}, 2018.

\bibitem{zhang2014facial}
Zhanpeng Zhang, Ping Luo, Chen~Change Loy, and Xiaoou Tang,
\newblock ``Facial landmark detection by deep multi-task learning,''
\newblock in {\em European Conference on Computer Vision}. Springer, 2014, pp.
  94--108.

\end{thebibliography}
}

\end{document}